# Generalized Learning with Rejection for Classification and Regression Problems

Amina Asif [a] and Fayyaz ul Amir Afsar Minhas [a,b,*]

[a] *Department of Computer and Information Sciences, Pakistan Institute of Engineering and Applied Sciences (PIEAS), PO Nilore, Islamabad, Pakistan.*
[b] *Department of Computer Science, University of Warwick, Coventry CV4 7AL, UK*

\* *corresponding author email:* fayyaz.minhas14@alumni.colostate.edu

## Abstract

Learning with rejection (LWR) allows development of machine learning systems with the ability to discard low confidence decisions generated by a prediction model. That is, just like human experts, LWR allows machine models to abstain from generating a prediction when reliability of the prediction is expected to be low. Several frameworks for this learning with rejection have been proposed in the literature. However, most of them work for classification problems only and regression with rejection has not been studied in much detail. In this work, we present a neural framework for LWR based on a generalized meta-loss function that involves simultaneous training of two neural network models: a predictor model for generating predictions and a rejecter model for deciding whether the prediction should be accepted or rejected. The proposed framework can be used for classification as well as regression and other related machine learning tasks. We have demonstrated the applicability and effectiveness of the method on synthetically generated data as well as benchmark datasets from UCI machine learning repository for both classification and regression problems. Despite being simpler in implementation, the proposed scheme for learning with rejection has shown to perform at par or better than previously proposed methods. Furthermore, we have applied the method to the problem of hurricane intensity prediction from satellite imagery. Significant improvement in performance as compared to conventional supervised methods shows the effectiveness of the proposed scheme in real-world regression problems.

# 1 Introduction

The primary goal in development of artificially intelligent systems is to achieve human-like decision making ability [1]–[5]. A human expert, when asked a yes/no question, would respond with either *yes*, *no* or *I do not know*. That is, when confused, a human abstains from making a decision, especially in high-risk situations like medicine. In conventional machine learning, a model is trained to produce some decision scores when given examples [6]–[12]. However, a typical machine learning system developed to answer such a yes/no or classification question will always produce a yes or no answer. That is, for a given input, a decision would always be produced, regardless of whether the model has been trained for handling such examples or not. This trait may not pose a big problem for low-risk systems like movie/song recommendation systems but can become undesirable for systems of sensitive nature like automated disease diagnostics, security threat detection or similar high-risk systems. For example, consider a scenario where some tests for diagnosing a life-threatening disease are conducted for a patient. A human doctor would not declare the patient positive or negative for the disease unless he/she is fairly sure. In case of any confusion, he/she would prescribe further tests before making the final decision. That is, instead of making a decision with low confidence, a human expert would abstain from making any decision at all. An artificially intelligent system, being expected to behave like humans, should have a similar characteristic. Recently, there has been a focus over developing machine learning models with this trait, i.e., development of *models that know what they do not know* [13]. The learning paradigm is known as *Learning with Rejection (LWR)* or *Learning with Abstention*.

To perform learning with rejection, several methods have been proposed that can determine reliability of predictions made by a machine learning model. For neural networks with *softmax* in the last layer, the simplest approach is the use of probability scores as confidence over reliability of predictions [14]. However, a high probability does not imply that the decision made by the classifier is correct due to problems in calibration of scores [15], [16]. To handle poor calibration, some methods for transformation of decision scores of models to confidence values have also been proposed [16]–[19]. Similarly, a confidence score based approach has been proposed for structured predictions in [20]. Ensembles of classifiers have also been used for estimation of confidence of predictions [21]. Another approach that uses agreement between a classifier and a modified nearest neighbor classifier's predictions, called the trust score, for confidence estimation has been proposed in [14]. Some of the methods in the literature comprise of classifiers with integrated option for abstention, the most recent being a neural network based method called *SelectiveNet* [22]–[25].

Another framework for Learning With Rejection based on Support Vector Machines (SVMs) has been proposed by Cortes et al. [26]. The idea is to develop a method that can accept or discard the classifier's predictions by learning two models, one to perform classification and the other to decide whether the classifier's decision should be accepted or not [26], [27]. Cortes et al. presented a hinge-loss based formulation for classification with rejection using Support Vector Machines (SVMs). A stochastic gradient descent based solution inspired from their formulation has been used for automated liver disease diagnosis by Hamid et al. in [28].

Existing methods in the literature for LWR mainly focus on classification tasks [27]. Regression using rejection has not been studied in much detail. In this work, we present a neural framework based on a generalized loss function with native support for a variety of machine learning tasks like classification, regression, etc. We have evaluated the performance of our method over synthetic and benchmark datasets for classification and regression. Furthermore, we have applied the proposed method over the problem of hurricane intensity estimation using satellite imagery. In section 2, mathematical formulation and experimental setup employed for evaluation of the method have been presented. Results have been reported and discussed in section 3 followed by conclusions in section 4.

## 2 Methods

In this section, we present the mathematical formulation of the proposed framework and details of the experimental setup employed for performance evaluation.

## 2.1 Mathematical Formulation

A typical supervised machine learning method takes $n$ training examples $x_1, x_2, \ldots, x_n$ and corresponding targets $y_1, y_2, \ldots, y_n$ to learn a prediction function $f(x; \theta)$ parameterized by learnable parameters $\theta$. In case of neural networks, $\theta$ correspond to the weights of the neural network. The prediction function is then used to generate decisions for novel examples. In conventional settings, $f(x; \theta)$ will produce a score for any feature vector which has the same dimensionality as the examples used for training regardless of whether it belongs to the same distribution as the training data or not. This characteristic is a potential source of poor performance since typical machine learning models do not give any direct information about the likeliness of the prediction being accurate. Our objective is to develop a method that, given an example, can tell whether the prediction by the machine learning model should be accepted or discarded. In line with the work of Cortes et al [26], we propose learning two neural networks, a prediction model $h(x; \boldsymbol{\theta}_h)$ and a rejection model $r(x; \boldsymbol{\theta}_r)$ for determining if the prediction should be accepted or rejected. $\boldsymbol{\theta}_h$ and $\boldsymbol{\theta}_r$ represent weights of the respective neural networks. The rejection model is trained to produce a positive score ($r(x) > 0$) for predictions that should be accepted and zero or negative ($r(x) \leq 0$) in case of rejections.

To develop prediction and rejection models for a machine learning task, we propose simultaneous training scheme for both models. During training, the goal is to learn $h(x)$ by minimizing a user-defined prediction error or loss function $l(h(x; \boldsymbol{\theta}_h), y)$ between prediction and target values over training examples. This can be any suitable loss function such as hinge loss for classification, square error loss for regression, etc. The rejection model $r(x)$ is trained to reject examples for which the prediction model $h(x)$ produces an error. This is achieved by designing a meta-loss function $L(x, y)$ that penalizes prediction errors by $h(x)$ as well as incorrect rejections by $r(x)$. More precisely, $L(x, y)$ should incur a penalty proportional to the prediction loss $l(h(x; \boldsymbol{\theta}_h), y)$ for cases in which rejection function accepts the prediction (i.e., $r(x) > 0$) and a penalty proportional to a rejection cost parameter $c > 0$ for those examples for which the prediction loss $l(h(x; \boldsymbol{\theta}_h), y)$ is sufficiently small but the rejection function rejects the prediction (i.e., $r(x) \leq 0$). We propose the following simple convex meta-loss function to achieve this behavior:

$$L(x, y) = \max\{0, \ r(x; \boldsymbol{\theta}_r) + l(h(x; \boldsymbol{\theta}_h), y), \ c(1 - r(x; \boldsymbol{\theta}_r))\}$$

It can be seen that, for a given example, $L(x, y)$ would produce a positive value if the loss $l(h(x; \boldsymbol{\theta}_h), y)$ for the example is positive and the example is not rejected, ($r(x) > 0$). A penalty of $c(1 - r(x; \boldsymbol{\theta}_r))$ is imposed in case of rejections. The hyper-parameter $c$ can be used to control the number of rejections by the model: a high value of $c$ implies high penalty over rejections during training and hence a lower number of examples would be rejected. On the other hand, a small value of $c$ favors a large number of rejections during training.

Given that $l(h(\boldsymbol{x}; \boldsymbol{\theta}_h), y)$ is convex, $L(\boldsymbol{x}, y)$ will be convex due to the convexity of the *max* operator [29]. Therefore, for convex prediction model loss functions, the proposed meta-loss being convex improves convergence of the models. We demonstrate the convexity of the proposed meta-loss function in Figure 1 which shows $L(\boldsymbol{x}, y)$ for $c = 2$. It can be seen that the minima lies where $h(\boldsymbol{x})$ produces zero error and the rejection score $r(\boldsymbol{x})$ is positive (no rejection). Thus, minimizing $L(\boldsymbol{x}, y)$ implies minimizing prediction model error $l(h(\boldsymbol{x}; \boldsymbol{\theta}_h), y)$ as well as the number of rejections.

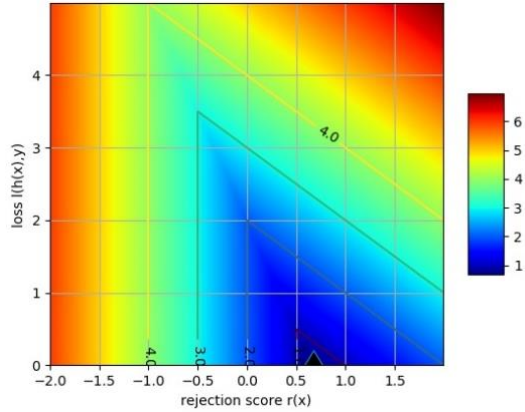

**Figure 1- Heat map of the proposed loss function. The black triangle represents the global minima. It can be seen that for a convex loss function *l*, the proposed meta-loss is also convex with a single global minimum.**

During training, the goal is to minimize the loss $L(\boldsymbol{x}, y)$ over all training examples. Therefore, the empirical error minimization can be expressed as:

$$\boldsymbol{\theta}_h^*, \boldsymbol{\theta}_r^* = argmin_{\boldsymbol{\theta}_h, \boldsymbol{\theta}_r} \sum_{i=1}^{n} L(\boldsymbol{x}_i, y_i).$$

During testing, examples are passed to both the models and only the predictions for which $r$ produces a positive score are accepted.

## 2.2 Experiments

To determine the effectiveness of our proposed scheme for regression and classification tasks, we evaluated the method over synthetic and real-world datasets for both classification and regression problems. We present details of each of the experiments in the following sections. To implement the classification regression and rejection models, we have used neural networks implemented using pyTorch [30]. Different architectures have been used for all the experiments. We present the details of the respective employed architectures in the following sections.

*Synthetic data*

To evaluate and analyze correctness of the proposed method for classification, we generated two groups of Gaussian distributed 2-dimensional data points with 200 examples each one centered at (-0.5, -0.5) and the other at (0.5,0.5). The first group is labeled as positive class and the other as negative class. The overlapping region between the two classes is the region of low confidence and hence test examples from this region should be rejected. Testing is performed over 200 examples of each class sampled from the same distribution as the training data.

To implement the classification model, we used a single layer neural network with one neuron with linear activation in the output layer. The rejection model has been implemented as a one hidden layer neural network. The hidden layer contains two neurons having *tanh* activations and the output layer consists of one neuron with linear activation. Hinge loss function has been used as the prediction model loss $l(h(x; \theta_h), y)$, in the proposed meta-loss. Performance evaluation has been performed by comparing AUC-ROC for predictions with and without rejections.

To test the performance of our method in regression settings using synthetic data, we used data from the same data distribution and the same architecture as described above. The only change was that we used epsilon-insensitive loss instead of hinge loss as the prediction model loss for training. Mean Squared Error and AUC-ROC have been used for performance evaluation.

*UCI datasets*

We have evaluated the performance of our proposed method on three datasets from UCI machine learning data repository, two for classification and one for regression [31]. For classification, we tested our method on Haberman's survival dataset and Australian credit approval dataset [32], [33]. Concrete strength dataset was used for evaluating our method's performance for regression [34].

Haberman's dataset consists of survival data of breast cancer patients who had undergone surgery. The dataset comprises of three features: patient's age, year of operation and number of axillary nodes. Labels have been assigned to examples on the basis of the duration of post-surgery survival, positive for the patients who survived five years or more and negative otherwise. For classification model, we used a single hidden layer neural network with 6 ReLU activated neurons in the hidden layer and one neuron with linear activation in the output layer. The rejection model consisted of two hidden layers, one with 32 and the other with 64 Tanh activated neurons. The output layer contained a single neuron with linear activation.

The Australian credit approval dataset is a binary classification dataset comprising of information about credit card applications. The dataset consists of 690 examples with 14 features. To model this problem, we used a single layer neural network with one linearly activated output neuron for classification. For rejection, a single hidden layer neural network has been used. The hidden layer contains 64 neurons with Tanh activation. The output layer contains one neuron with linear activation.

For both Haberman's and Australian credit datasets, we have compared the performance of our method with the SVM based learning with rejection proposed by Cortes et al [26]. 5-fold cross-validation has been performed for a fair comparison. Classification error has been used as the performance evaluation metric in this case.

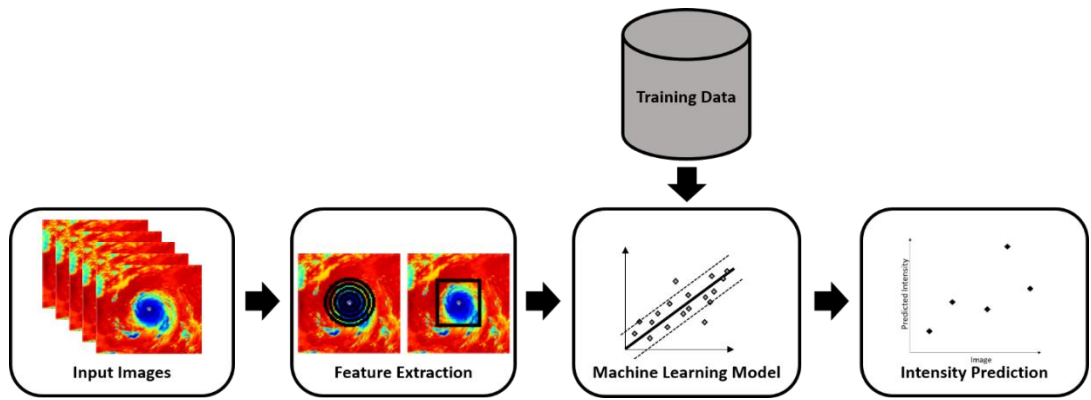

**Figure 2- Illustration of workflow of PHURIE [35].**

To evaluate the performance of our method in regression settings, we tested it on concrete compressive strength dataset from UCI repository. The dataset consists of 1030 examples with 8 features. The task is to predict the compressive strength given features of concrete representing its ingredients and age. As per the proposed approach, to model this problem, we used one neural network for performing regression and the other for rejection. The regression network consists of two hidden layers, one with 64 neurons and the other with 32 neurons. ReLU activation is applied over both the hidden layers. The output layer comprises of a single linearly activated neuron. To implement rejection model, we used a similar architecture i.e., two hidden layers, one containing 64 neurons and the other 32 neurons. Tanh activation is used in both the layers. The output layer contains one neuron with linear activation. During evaluation, we used randomly chosen 90% examples for training and the rest for testing. The experiment has been repeated 20 times. We used Mean Squared Error with and without rejection for performance evaluation. We compared the performance of our method with SelectiveNet proposed by Geifman et al. [25] through the same evaluation protocol.

*Hurricane Intensity Prediction*

In our previous work, we developed a system named *PHURIE* for predicting hurricane intensities in knots using satellite infrared images [35]. The system takes as input infrared images, computes features signifying cloud structure in the image and uses a trained Support Vector Regression (SVR) model for intensity prediction. The features include mean, minimum, maximum, standard deviation and entropy of bands around the center of the hurricane. Additionally, variance of deviation angle histogram of an image is also used as a feature [36]. Workflow of PHURIE is shown in Figure 2. While developing PHURIE, we compared the performance of different regression methods for the proposed set of features including Ordinary Least Square (OLS) regression, SVR, neural networks and XGBoost and SVR with RBF kernel outperformed other methods and was therefore chosen for training the final model for the system to be deployed [37].

PHURIE was built using conventional supervised learning approach and therefore, does not support learning with rejection. In this experiment, we applied the proposed approach for learning with rejection using data and features used in PHURIE. As per the proposed approach, we needed two neural network models, one to perform regression and the other for rejection. The network architecture used for regression has been chosen such that it performs at par with RBF-SVR based PHURIE model. To mimic SVR, we used a single layer neural network with one output neuron and linear activation. Epsilon-insensitive loss function was used for evaluating regression error. The SVR model in PHURIE used RBF kernel. To get similar performance using a single layer neural network we first applied RBF kernel approximation [38] over the features and used transformed examples as inputs to the network. The rejection network is also a single layer network with one output neuron and linear activation.

We have performed Leave One Year Out (LOYO) cross-validation over data from years 2004-2009 for performance evaluation. In LOYO cross-validation, we remove one year's examples from the data, use the remaining set for training and test over the left-out year's data. Root Mean Squared Error (RMSE) has been used as the performance evaluation metric.

## 3 Results and Discussion

In this section, we present and discuss results obtained for experiments described in the previous section.

### 3.1 Synthetic Data

The decision boundaries produced by the classifier and the rejection function for synthetic data are presented in Figure 3. We obtained an AUC-ROC of 0.84 for classification without rejection for the classifier. Figure 3 shows the decision boundary learnt for both the classification and rejection functions. It can be seen that the rejection function encloses the region of overlap between positive and negative classes i.e., the region of low confidence in this case. AUC-ROC by removing predictions for examples from this region increases to 0.88. In Figure 4 we present a plot depicting the trend in AUC-ROC if we reject different fractions of test data using the rejection model. The rejections are performed by removing top *n* predictions according to the sorted list of rejection scores. We compare the AUCs with those of obtained by performing same number of random rejections. It can be seen that there is a consistently increasing trend in AUCs for rejections using the learnt rejection model as compared to the random rejections (for which the AUC remains almost constant), hence demonstrating the effectiveness of the proposed scheme.

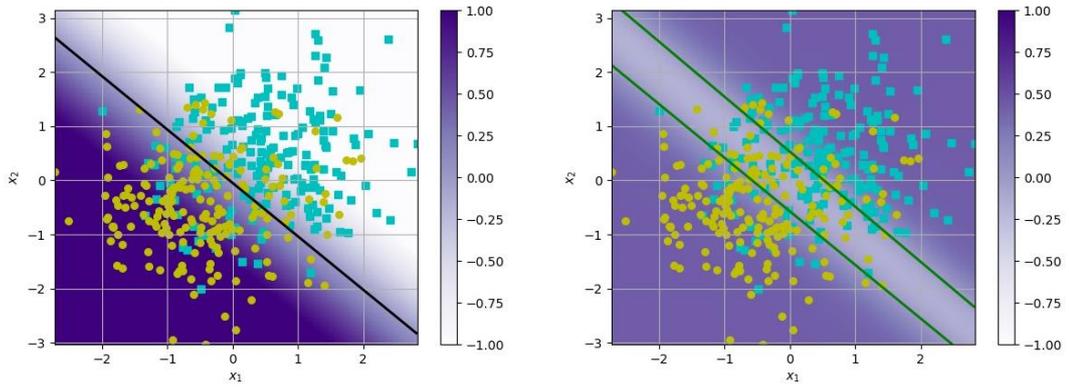

**Figure 3- Boundaries obtained for synthetic binary classification data for (a) classification model and (b) rejection model.**

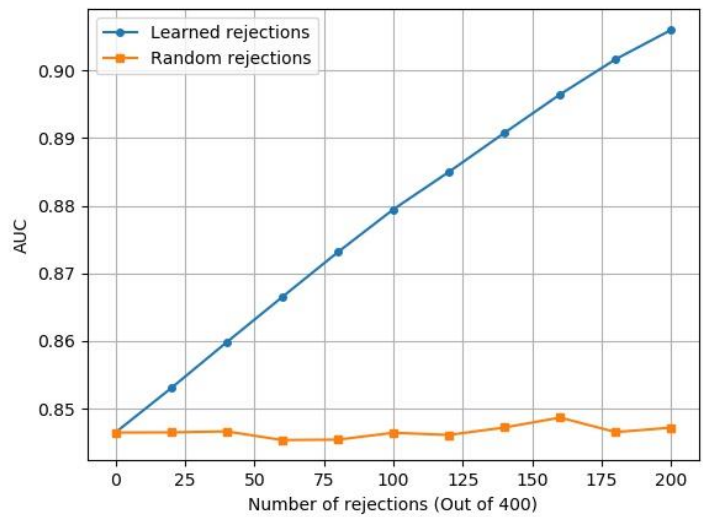

**Figure 4- Comparison between AUC after performing random rejections and rejecting predictions using the proposed scheme.**

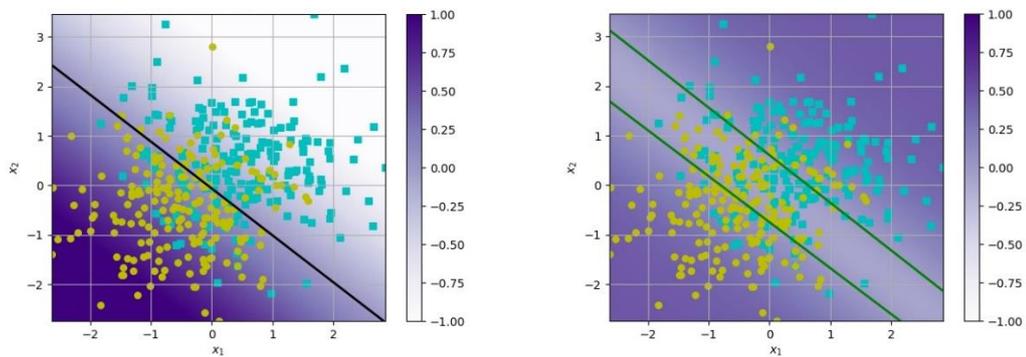

**Figure 5- Boundaries produced by (a) prediction model trained using regression loss and (b) rejection model for synthetic data.**

**Table 1- Classification error for classification datasets from UCI repository.**

| Dataset | Fraction rejected | DHL [26] | CHR [26] | Our method | |
|---|---|---|---|---|---|
| | | | | Without rejection | With rejection |
| **Australian** | 0.17 | 0.35±0.10 | 0.07±0.02 | 0.12±0.03 | 0.08±0.02 |
| **Haberman** | 0.44 | 0.25±0.11 | 0.10±0.05 | 0.22±0.03 | 0.13±0.05 |

In addition to solving the synthetic classification problem using the classification loss, we modeled the same problem using a regression loss for prediction model as well. The boundary produced by the regression model and the rejection model are presented in Figure 5. Here again, the AUC-ROC without rejection is 0.85 and by removing predictions for which the rejection function produces a negative score, the AUC increases to 0.9. The MSE without rejection was 0.66 which reduced to 0.51 after removing predictions that were rejected.

## 3.2 UCI datasets

In this section, we present the results proposed using the proposed scheme for the three UCI repository datasets. We evaluated the performance of our technique on three UCI datasets: two for classification and one for regression.

For classification datasets, Haberman's survival and Austalian credit data, the results for 10 runs of 5-fold cross-validation are presented in Table 1. We have compared our results with SVM based LWR proposed by Cortes et al. [26]. We reject the same fraction as other methods by removing the fraction of examples producing lowest scores when passed through the rejection model. It can be seen that the classification error of our method, 0.08 for Australian and 0.13 for Haberman's, is comparable to the SVM based LWR (0.07 and 0.10) proving our technique to be as effective as theirs for classification problems.

The results obtained over the regression dataset of Concrete compressive strength are presented in Table 2. We compare the performance of our method with SelectiveNet [25]. We present Mean Squared Error values for different fractions of rejections. It can be seen that our method produces better mean MSE values across all fractions of rejections, therefore proving that effectiveness of the proposed scheme.

## 3.3 Hurricane Intensity prediction: PHURIE with rejection

The results for Leave One Year Out cross-validation for the problem of hurricane intensity prediction using satellite imagery are given in Table 3. We present a comparison among results obtained for different fractions of rejection. It can be seen that there is a consistent downwards trend in the RMSEs with increasing fractions of rejections. In Figure 6 we present plots comparing RMSEs using our method for performing rejections and random rejections for years 2004 and 2005. As evident from the plots, the rejection model learns to reject examples such that the overall performance of the system improves i.e., the rejection model successfully identifies low confidence regions of the feature space.

**Table 2- Mean Squared Error values for Concrete Strength dataset.**

| Fraction rejected | SelectiveNet [25] | Our Method |
|---|---|---|
| **0** | 38.45 | **33.4** |
| **0.10** | 35.35 | **30.3** |
| **0.20** | 30.48 | **28.3** |
| **0.30** | 27.94 | **26.3** |
| **0.40** | 27.12 | **24.6** |
| **0.50** | 26.81 | **24.0** |

To further demonstrate the effectiveness of the proposed scheme, we present images for the top nine rejected examples from year 2004 in Figure 7**.** It can be seen that the predictions for the rejected images have high RMSE. That is, the rejection model has successfully learned to reject examples for which the error would be high. It can be seen that the rejected images are either noisy or have ill-formed cloud structures. Therefore, the rejection model trained using the proposed strategy has successfully achieved the ability to anticipate which predictions would produce high errors and can reject such predictions, hence, leading to better performance of the system.

**Table 3- RMSE with and without rejection for hurricane intensity estimation using satellite images.**

| Year | Proposed Method RMSE in knots for different fractions of rejection | | | | | |
|---|---|---|---|---|---|---|
| | **0.0** | **0.1** | **0.2** | **0.3** | **0.4** | **0.5** |
| **2004** | 14.5 | 13.5 | 12.3 | 11.3 | 10.9 | 10.6 |
| **2005** | 12.5 | 11.2 | 10.8 | 10.3 | 10.1 | 9.8 |
| **2006** | 13.4 | 11.7 | 11.3 | 10.7 | 10.5 | 10.2 |
| **2007** | 10.9 | 10.3 | 9.7 | 9.4 | 9.2 | 8.5 |
| **2008** | 14.0 | 13.4 | 13.2 | 13.0 | 12.8 | 13.1 |
| **2009** | 14.1 | 13.9 | 13.2 | 12.7 | 12.3 | 12.0 |
| **Mean** | 13.2 | 12.3 | 11.7 | 11.2 | 10.9 | 10.7 |

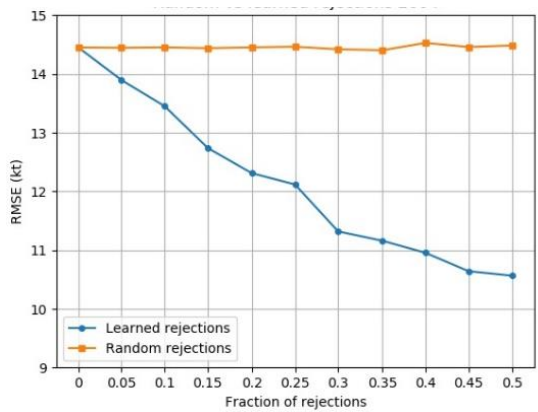
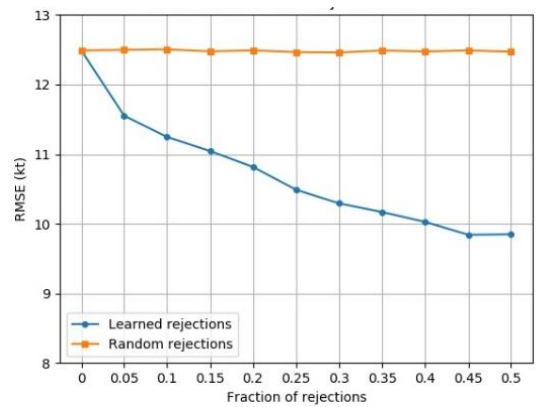

(a)          (b)

**Figure 6- Comparison between random and learnt rejections using the proposed scheme for years (a) 2004 and (b) 2005. It can be seen that with increasing fraction of rejections, our method produces consistent decrease in RMSE.**

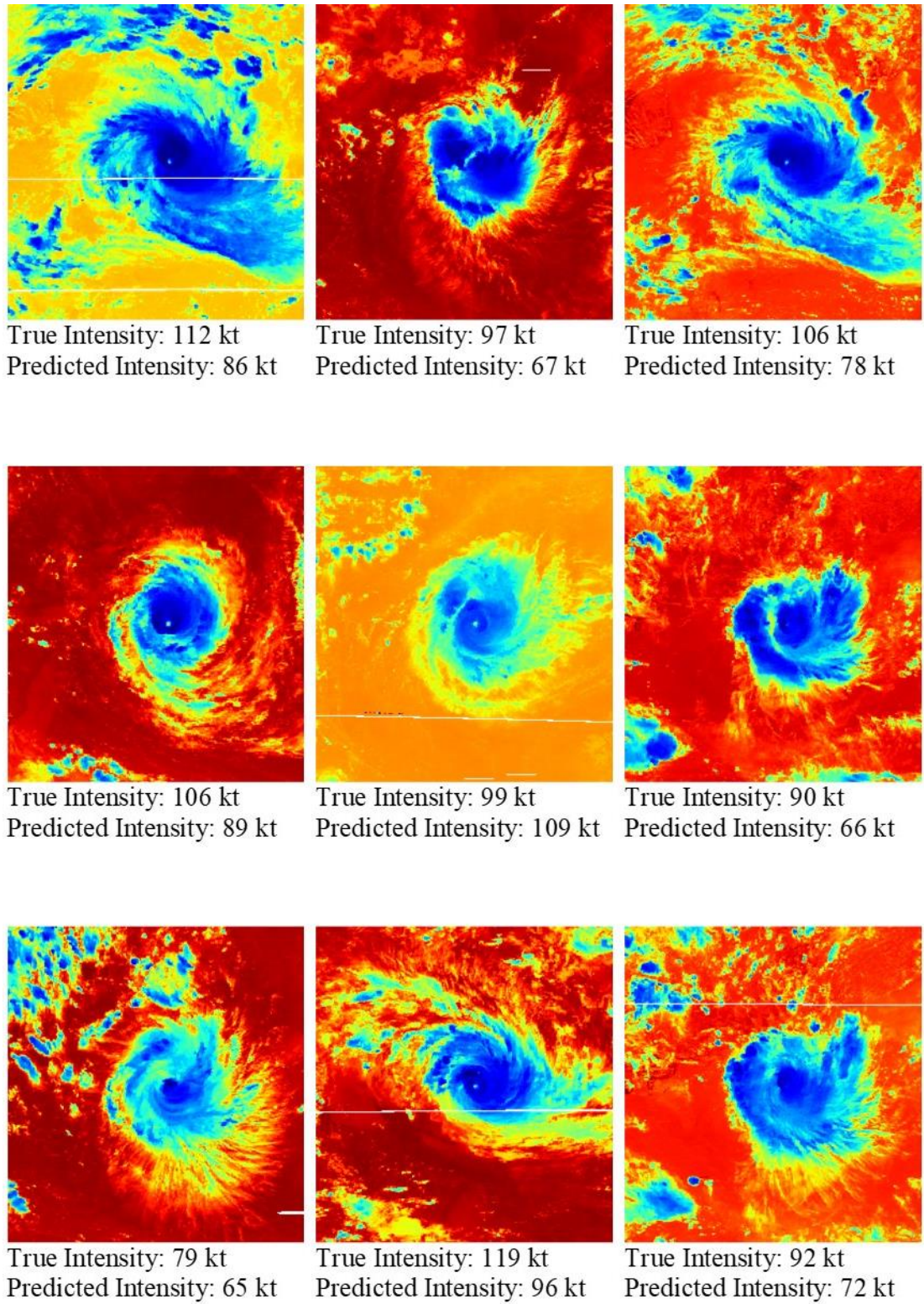

**Figure 7- Top rejected images for year 2004. It can be seen that the RMSE of predictions for rejected images is high, i.e., the rejection model is correctly identifying the predictions in which error is expected to be high. The high RMSE in these images can be attributed to noise or the ill-formed cloud structure.**

## 4 Conclusions

In this study, we presented a generalized scheme for learning with rejection, i.e., a method using which predictions that are expected to be inaccurate can be rejected. Our technique learns two models simultaneously, one for learning task (classification or regression) and the other to perform rejections. We have proposed a meta-loss function to be used for training of the two models that attempts to minimize the training error as well as the number of rejections. Using appropriate value of cost of rejection, it favors rejection of an example over a large training error. Given a convex loss function for the learning model, our proposed meta-loss is also convex, hence improving the chances of models' convergence. We have demonstrated the applicability and effectiveness of the proposed method in both classification and regression problems. We have performed experiments over synthetic data and UCI repository datasets. Furthermore, we have applied the proposed method on the problem of hurricane intensity estimation using satellite imagery. Results have shown that predictions rejected using our method have improved the overall performance of the systems in all the experiments. Moreover, the generalized nature of our formulation allows it to be equally effective in classification and regression problems. Also, learning two models, one for learning and the other for rejection allows for using models of different complexities for the two tasks.

# References

[1] L. Deng, "Artificial Intelligence in the Rising Wave of Deep Learning: The Historical Path and Future Outlook [Perspectives]," *IEEE Signal Process. Mag.*, vol. 35, no. 1, pp. 180–177, Jan. 2018.

[2] H. Lu, Y. Li, M. Chen, H. Kim, and S. Serikawa, "Brain intelligence: go beyond artificial intelligence," *Mob. Netw. Appl.*, vol. 23, no. 2, pp. 368–375, 2018.

[3] G. Gurkaynak, I. Yilmaz, and G. Haksever, "Stifling artificial intelligence: Human perils," *Comput. Law Secur. Rev.*, vol. 32, no. 5, pp. 749–758, Oct. 2016.

[4] M. O. Riedl and B. Harrison, "Using stories to teach human values to artificial agents," in *Workshops at the Thirtieth AAAI Conference on Artificial Intelligence*, 2016.

[5] Y. Zang, F. Zhang, C. Di, and D. Zhu, "Advances of flexible pressure sensors toward artificial intelligence and health care applications," *Mater. Horiz.*, vol. 2, no. 2, pp. 140–156, 2015.

[6] S. B. Kotsiantis, I. Zaharakis, and P. Pintelas, *Supervised machine learning: A review of classification techniques*. 2007.

[7] H. Wang, N. Wang, and D.-Y. Yeung, "Collaborative deep learning for recommender systems," in *Proceedings of the 21th ACM SIGKDD international conference on knowledge discovery and data mining*, 2015, pp. 1235–1244.

[8] C. Biancalana, F. Gasparetti, A. Micarelli, A. Miola, and G. Sansonetti, "Context-aware movie recommendation based on signal processing and machine learning," in *Proceedings of the 2nd Challenge on Context-Aware Movie Recommendation*, 2011, pp. 5–10.

[9] X. Zheng, Z. Zeng, Z. Chen, Y. Yu, and C. Rong, "Detecting spammers on social networks," *Neurocomputing*, vol. 159, pp. 27–34, 2015.

[10] D. Ruths and J. Pfeffer, "Social media for large studies of behavior," *Science*, vol. 346, no. 6213, pp. 1063–1064, Nov. 2014.

[11] K. Kourou, T. P. Exarchos, K. P. Exarchos, M. V. Karamouzis, and D. I. Fotiadis, "Machine learning applications in cancer prognosis and prediction," *Comput. Struct. Biotechnol. J.*, vol. 13, pp. 8–17, 2015.

[12] M. W. Libbrecht and W. S. Noble, "Machine learning applications in genetics and genomics," *Nat. Rev. Genet.*, vol. 16, no. 6, p. 321, 2015.

[13] D. Madras, T. Pitassi, and R. Zemel, "Predict responsibly: Increasing fairness by learning to defer," 2018.

[14] H. Jiang, B. Kim, M. Guan, and M. Gupta, "To Trust Or Not To Trust A Classifier," in *Advances in Neural Information Processing Systems 31*, S. Bengio, H. Wallach, H. Larochelle, K. Grauman, N. Cesa-Bianchi, and R. Garnett, Eds. Curran Associates, Inc., 2018, pp. 5541–5552.

[15] A. Kendall and Y. Gal, "What uncertainties do we need in bayesian deep learning for computer vision?," in *Advances in neural information processing systems*, 2017, pp. 5574–5584.

[16] C. Guo, G. Pleiss, Y. Sun, and K. Q. Weinberger, "On calibration of modern neural networks," in *Proceedings of the 34th International Conference on Machine Learning-Volume 70*, 2017, pp. 1321–1330.

[17] A. Niculescu-Mizil and R. Caruana, "Predicting good probabilities with supervised learning," in *Proceedings of the 22nd international conference on Machine learning*, 2005, pp. 625–632.
[18] J. Platt and others, "Probabilistic outputs for support vector machines and comparisons to regularized likelihood methods," *Adv. Large Margin Classif.*, vol. 10, no. 3, pp. 61–74, 1999.
[19] B. Zadrozny and C. Elkan, "Transforming classifier scores into accurate multiclass probability estimates," in *Proceedings of the eighth ACM SIGKDD international conference on Knowledge discovery and data mining*, 2002, pp. 694–699.
[20] V. Kuleshov and P. S. Liang, "Calibrated structured prediction," in *Advances in Neural Information Processing Systems*, 2015, pp. 3474–3482.
[21] B. Lakshminarayanan, A. Pritzel, and C. Blundell, "Simple and scalable predictive uncertainty estimation using deep ensembles," in *Advances in Neural Information Processing Systems*, 2017, pp. 6402–6413.
[22] P. L. Bartlett and M. H. Wegkamp, "Classification with a reject option using a hinge loss," *J. Mach. Learn. Res.*, vol. 9, no. Aug, pp. 1823–1840, 2008.
[23] M. Yuan and M. Wegkamp, "Classification methods with reject option based on convex risk minimization," *J. Mach. Learn. Res.*, vol. 11, no. Jan, pp. 111–130, 2010.
[24] G. Fumera and F. Roli, "Support vector machines with embedded reject option," in *International Workshop on Support Vector Machines*, 2002, pp. 68–82.
[25] Y. Geifman and R. El-Yaniv, "SelectiveNet: A Deep Neural Network with an Integrated Reject Option," *ArXiv Prepr. ArXiv190109192*, 2019.
[26] C. Cortes, G. DeSalvo, and M. Mohri, "Learning with rejection," in *International Conference on Algorithmic Learning Theory*, 2016, pp. 67–82.
[27] B. Hanczar, "Performance visualization spaces for classification with rejection option," *Pattern Recognit.*, p. 106984, Jul. 2019.
[28] K. Hamid, A. Asif, W. Abbasi, D. Sabih, and F. U. A. A. Minhas, "Machine Learning with Abstention for Automated Liver Disease Diagnosis," in *2017 International Conference on Frontiers of Information Technology (FIT)*, 2017, pp. 356–361.
[29] M. C. Grant and S. P. Boyd, "Graph implementations for nonsmooth convex programs," in *Recent advances in learning and control*, Springer, 2008, pp. 95–110.
[30] N. Ketkar, "Introduction to pytorch," in *Deep learning with python*, Springer, 2017, pp. 195–208.
[31] C. L. Blake and C. J. Merz, *UCI repository of machine learning databases, 1998*. 1998.
[32] S. J. Haberman, "Generalized residuals for log-linear models," in *Proceedings of the 9th international biometrics conference*, 1976, pp. 104–122.
[33] J. R. Quinlan, "Simplifying decision trees," *Int. J. Man-Mach. Stud.*, vol. 27, no. 3, pp. 221–234, 1987.
[34] I.-C. Yeh, "Modeling of strength of high-performance concrete using artificial neural networks," *Cem. Concr. Res.*, vol. 28, no. 12, pp. 1797–1808, 1998.
[35] A. Asif, M. Dawood, B. Jan, J. Khurshid, M. DeMaria, and F. ul A. A. Minhas, "PHURIE: Hurricane Intensity Estimation from Infrared Satellite Imagery using Machine Learning."
[36] E. A. Ritchie, K. M. Wood, O. G. Rodríguez-Herrera, M. F. Piñeros, and J. S. Tyo, "Satellite-Derived Tropical Cyclone Intensity in the North Pacific Ocean

Using the Deviation-Angle Variance Technique," *Weather Forecast.*, vol. 29, no. 3, pp. 505–516, Dec. 2013.

[37]   D. Basak, S. Pal, and D. C. Patranabis, "Support vector regression," *Neural Inf. Process.-Lett. Rev.*, vol. 11, no. 10, pp. 203–224, 2007.

[38]   A. Rahimi and B. Recht, "Weighted sums of random kitchen sinks: Replacing minimization with randomization in learning," in *Advances in neural information processing systems*, 2009, pp. 1313–1320.